\documentclass[a4paper, twoside, 11pt]{report}

\usepackage[english]{babel}
\usepackage[utf8x]{inputenc}
\usepackage[T1]{fontenc}
\usepackage{listings}
\usepackage{hyperref}
\hypersetup{colorlinks=false}
\usepackage{lscape}
\usepackage{subfigure}
\usepackage{amsmath}
\usepackage{amssymb}
\DeclareMathOperator{\E}{\mathbb{E}}
\usepackage{float}
\usepackage{commath}
\usepackage{graphicx}
\usepackage{hyperref}
\usepackage{algorithm}
\usepackage[noend]{algpseudocode}
\usepackage[colorinlistoftodos]{todonotes}

\usepackage[a4paper,top=3cm,bottom=2cm,left=3cm,right=3cm,marginparwidth=1.75cm]{geometry}

\title{Emotion Generation and Recognition: A StarGAN Approach}
\author{Aritra Banerjee\\CID: 01539378}

\begin{document}
\begin{titlepage}

\newcommand{\HRule}{\rule{\linewidth}{0.5mm}} 


\includegraphics[width=8cm]{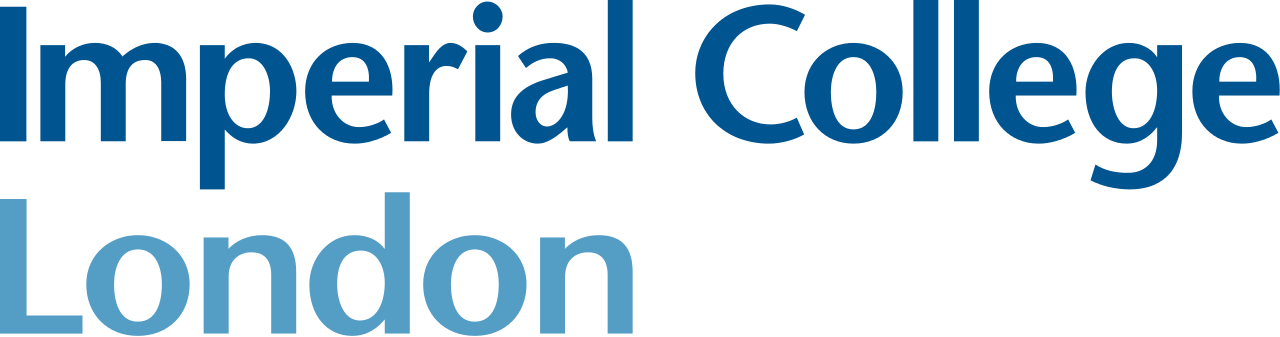}\\[1cm] 
 

\center 

\quad\\[1.5cm]
\textsc{\Large Imperial College London}\\[0.5cm] 
\textsc{\large Department of Computing}\\[0.5cm] 
\textsc{\large Independent Study Option (CO-512)}\\[0.5cm]
\makeatletter
\HRule \\[0.4cm]
{ \huge \bfseries \@title}\\[0.4cm] 
\HRule \\[1.5cm]
 

\begin{minipage}{0.4\textwidth}
\begin{flushleft} \large
\emph{Author:}\\
\@author 
\end{flushleft}
\end{minipage}
~
\begin{minipage}{0.4\textwidth}
\begin{flushright} \large
\emph{Supervisor:} \\
Dimitrios Kollias 
\end{flushright}
\end{minipage}\\[3cm]
\makeatother


{\large A report submitted for fulfillment of the degree}\\[0.5cm]
{\large \emph{MSc Computing (Machine Learning)}}\\[0.5cm]
{\large \today}\\[2cm] 

\vfill 

\end{titlepage}

\begin{abstract}
The main idea of this ISO is to use StarGAN(A type of GAN model) to perform training and testing on an emotion dataset resulting in a emotion recognition which can be generated by the valence arousal score of the 7 basic expressions. We have created an entirely new dataset consisting of 4K videos.\\
This dataset consists of all the basic 7 types of emotions:
\begin{itemize}
    \item Happy
    \item Sad
    \item Angry
    \item Surprised
    \item Fear
    \item Disgust
    \item Neutral
\end{itemize}
We have performed face detection and alignment followed by annotating basic valence arousal values to the frames/images in the dataset depending on the emotions manually. Then the existing StarGAN model is trained on our created dataset after which some manual subjects were chosen to test the efficiency of the trained StarGAN model. 
\end{abstract}

\renewcommand{\abstractname}{Acknowledgements}
\begin{abstract}
I would like to thank everyone who have provided particularly useful assistance, technical or otherwise, during my Independent Study Option (ISO). I would like to thank my supervisor Dimitrios Kollias specifically for being an outstanding mentor and guide throughout the entire project. He guided me through every step and ensured that I work on this ISO according to my full potential within the given deadline.
\end{abstract}

\tableofcontents
\listoffigures

\chapter{Introduction}
We will be using a type of Generative Adversarial Network (GAN) \cite{goodfellow2014generative}: StarGAN \cite{choi2018stargan} in this project for emotion recognition using valence arousal scores. It is basically a multi-task GAN which uses the Generator to produce fake images and the discriminator to identify the real and fake images along with emotion recognition on the basis of attributes, i.e. valence arousal score in our ISO. Applications of emotion recognition are in many fields such as the medical \cite{tagaris1,tagaris2,kollias13}, marketing etc. 

\section{Generative Adversarial Networks(GAN)}
A GAN is form of unsupervised learning that simultaneously trains two models: a generative model G that captures the data distribution by using a latent noise \cite{raftopoulos2018beneficial} vector, and a discriminative model D that estimates the probability that a sample came from the training data rather than G \cite{goodfellow2014generative}.The training procedure is such that the G maximizes the probability of D making a mistake. This framework corresponds to a minimax two-player game. The two networks basically contests with each other in a zero-sum game framework.\\
The D is trained to maximize the probability of assigning the correct label to both training examples and samples from G.
We train D to maximize the probability of assigning the
correct label to both training examples and samples from G. We simultaneously train G to minimize $\log(1 − D(G(z)))$.
So, basically, the D and G minimax equation with value function V (G;D) can be written as:
\begin{equation}
    \min_{G} \max_{D} V(D,G) =  \E_{x\sim p_{data}(x)}[\log(D(x))] + \E_{z\sim p(z)}[\log(1-D(G(z)))]
\end{equation}
Usually a model of GAN looks like this:
\begin{figure}[H]
  \centering
  \includegraphics[width=0.60\textwidth]{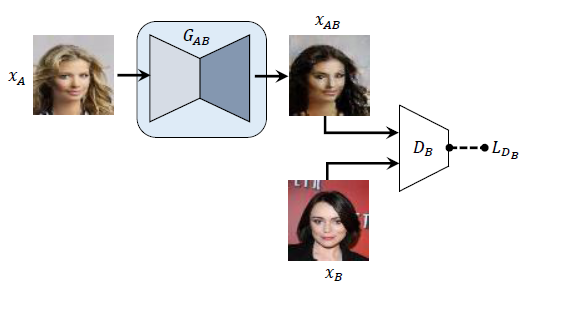}
  \caption{General Overview of a GAN consisting of Discriminator(D) and Generator(G)}
  \label{fig:gan}
\end{figure}
In adversarial nets framework, the generative model is put up against an adversary: a discriminative model that learns and determines whether a sample is from the model distribution or the
data distribution. The generative model can be thought of as parallel to a team of money launderers, trying to produce fake money and use it without detection, while the discriminative model is
parallel to the police, trying to detect the fake currency. Competition in this game drives both teams to improve their methods until the fake ones are indistinguishable from the real articles.
\section{StarGAN}
The StarGAN \cite{choi2018stargan} is basically a type of GAN which solves the problem of multi-domain image to image translation. The existing approaches reduces the robustness of a problem when used for multi-domains. So, basically for each pair of image domain we do not have to create a different network.
The task of image-to-image translation is to change a
given aspect of a given image to another,i.e, changing
the facial expression of a person from neutral to happy.\\
The emotion dataset we have created has 7 different basic expressions which are the 7 labels which we will use \cite{kollias11,kollias12}.\\
The basic structure of the StarGAN can be shown as below:
\begin{figure}[H]
  \centering
  \includegraphics[width=0.65\textwidth]{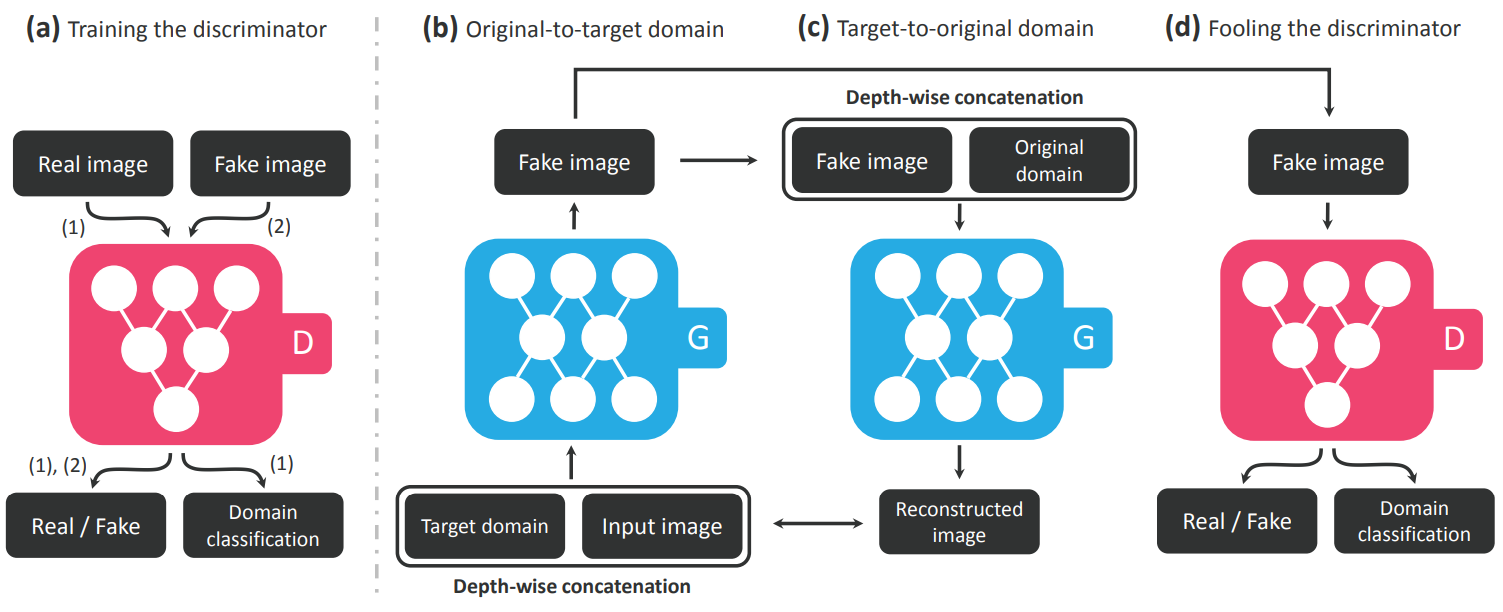}
  \caption{General Overview of a StarGAN consisting of Discriminator(D) and Generator(G)}
  \label{fig:stargan}
\end{figure}
As \textbf{attribute} we have annotated the videos in the dataset based on their valence and arousal score.\\
As \textbf{domain} we have the different videos pertaining to the same emotion which shares the same attribute value i.e., angry, sad etc.\\
\\
\textbf{Conditional} GAN-based image generation have been actively studied.
Prior studies have provided both the discriminator and generator with class information in order to generate samples conditioned on the class\cite{mirza2014conditional},\cite{odena2016semi}.This idea of conditional image generation has been successfully applied to domain transfer\cite{kim2017learning} which is a building block for this project.\\
Recent works have achieved significant results in the area of image-to-image translation\cite{kim2017learning},\cite{isola2017image}. For instance, pix2pix \cite{isola2017image} learns this task in a supervised manner using conditional GANs\cite{mirza2014conditional}. It combines an adversarial loss with an L1 penalty loss, thus requires paired data samples.\\
\\
The basic idea in the Star GAN approach is to train a single generator G that learns mappings among multiple domains, in our case it is 7 basic expressions. To reach this, we need to train G
to translate \cite{goudelis2013exploring} an input image x to an output image y conditioned
on the target domain label $c$, $G(x,c) \longrightarrow y$. 
We randomly generate the target domain label $c$ ,i.e. from valence or arousal score so that the generator learns to flexibly translate the input image. An auxiliary classifier\cite{odena2017conditional} was introduced in the Star GAN approach that allows a single discriminator to control multiple domains. So, the discriminator produces probability distributions over both source labels and domain labels, $D: x \longrightarrow {D_{src}(x),D_{cls}(x)}$.
\chapter{Background}
In this chapter we compare the StarGAN against recent facial expressions studies conducted by different approaches. These are basically the baseline models which perform image-to-image translation \cite{kollias8,kollias9} for different domains and how the Star GAN has improved upon these models.
We explain four different baseline models here in this case:
\\
\begin{itemize}
    \item \textbf{DIAT}\cite{li2016deep} or Deep identity-aware transfer uses an adversarial loss to learn the mapping from i $\in$ I to j $\in$ J, where i and j are face images in two different domains I and J, respectively.
    This method adds a regularization term on the mapping as $\|i- F(G(i))\|_{1}$ to preserve identity features of the source image, where F is a feature extractor pretrained on a face recognition task.
    \begin{figure}[H]
    \centering
    \includegraphics[width=0.95\textwidth]{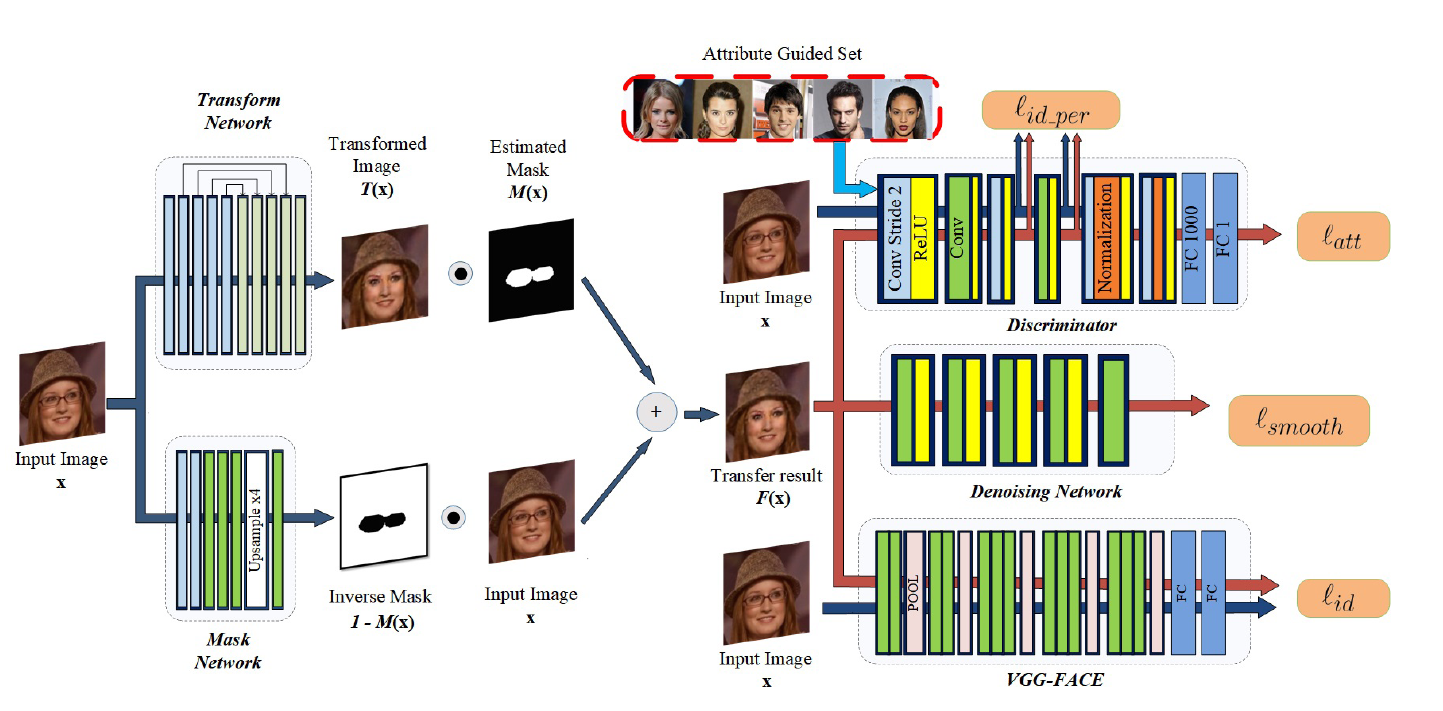}
    \caption{General Overview of a DIAT model}
    \label{fig:diat}
    \end{figure}
    \item \textbf{CycleGAN}\cite{zhu2017unpaired} uses an adversarial loss to learn the mapping between two different domains I and J. This method regularizes the mapping via cycle consistency losses, $\|i-(G_{JI}(G_{IJ}(i)))\|_{1}$ and $\|j-(G_{IJ}(G_{JI}(j)))\|_{1}$.
    This method requires \textbf{'n'} generators and discriminators for each pair of \textbf{'n'} different domains. So, in our case we would need 7 different generators and discriminators.
    \item \textbf{IcGAN}\cite{perarnau2016invertible} or Invertible Conditional GAN basically combines an encoder with a cGAN \cite{mirza2014conditional} model. cGAN learns the mapping $G :$ \textit{\{z,c\}} $\longrightarrow i$ that generates an image i conditioned on both the latent vector z and the conditional vector c. On top of that the IcGAN introduces the encoder to learn the inverse mappings of cGAN, $E_{z}:i\longrightarrow z$ and $E_{c}:i\longrightarrow c$. This allows IcGAN to synthesis images by only changing the conditional vector and preserving the latent vector.
    \begin{figure}[H]
    \centering
    \includegraphics[width=0.95\textwidth]{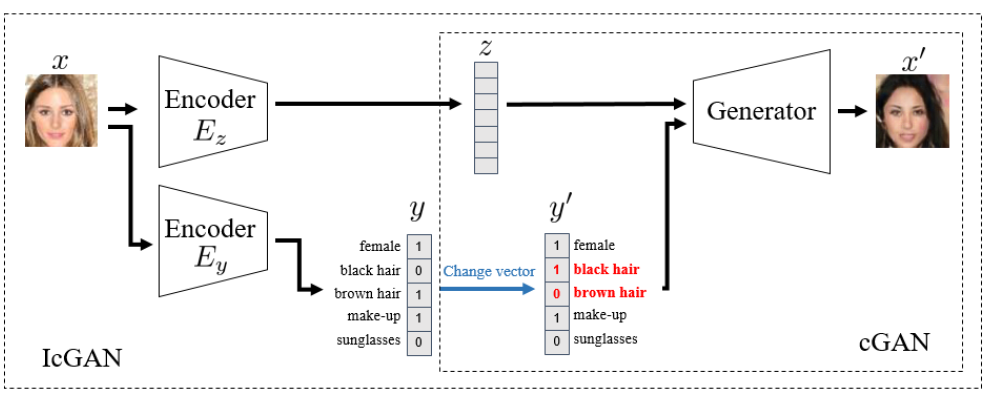}
    \caption{An IcGAN model consisting of Encoder and a conditional GAN generator}
    \label{fig:icgan}
    \end{figure}
    \item \textbf{DiscoGAN}\cite{kim2017learning} is the basic foundation baseline of StarGAN as Disco GAN introduces cross relation among different domains. The Disco GAN solves the relations between different unpaired domains with different label values for each domain.
    \begin{figure}[H]
    \centering
    \includegraphics[width=0.95\textwidth]{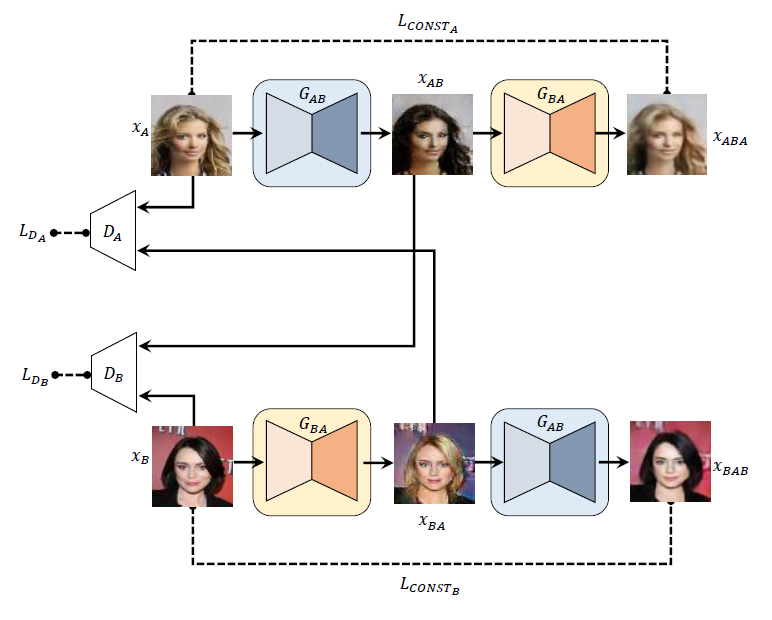}
    \caption{General Overview of a DiscoGAN consisting of Discriminator(D) and Generator(G)}
    \label{fig:discogan}
    \end{figure}
\end{itemize}
\chapter{Collecting and Preprocessing the Dataset}
In this chapter we specifically elaborate how the dataset was collected manually from internet sources as in \cite{kollias15}. Following which face detection was performed on the videos and alignment on the faces was performed so it is easier for the StarGAN to train which resulted in individual frames/images. After that each frame/image was annotated based on its valence arousal score \cite{kollias4,kollias5}.\\
Overall around 484 vidoes were collected spanning more or less equally among the 7 different expressions which resulted in  $\sim$250K frames/images.
\section{Collection of Dataset}
We used the website \href{https://www.shutterstock.com/}{https://www.shutterstock.com/} to collect the videos manually that was required for this ISO.\\
The following expressions were elaborately searched along with \cite{doulamis1999interactive,simou2008image,simou2007fire} its synonymous words throughout the website and the resulting videos divided into 7 different folders:\\
\\
\textbf{happy} : glad, pleased, delighted, cheerful, ecstatic, joyful, thrilled, upbeat, overjoyed, excited, amused, astonished.\\
\textbf{neutral} : serene, calm, at ease, relaxed, inactive, indifferent, cool, uninvolved.\\
\textbf{angry} : enraged, annoyed, frustrated, furious, heated, irritated, outraged, resentful, fierce, hateful, ill-tempered, mad, infuriated, wrathful.\\
\textbf{disgust} : antipathy, dislike, loathing, sickness.\\
\textbf{fear} : horror, terror, scare, panic, nightmare, phobia, tremor, fright, creeps, dread, jitter , cold feet, cold sweet.\\
\textbf{sad} : depressed, miserable, sorry, pessimistic, bitter, heartbroken, mournful, melancholy, sorrowful, down, blue, gloomy.\\
\textbf{surprise} : awe, amazement, curious, revelation, precipitation, suddenness, unforeseen, unexpected.\\
\\
Using these searches we developed the dataset folder and hence the number of videos/identities per emotion is :\\
\begin{itemize}
\item Happy - 76 videos.
\item Neutral - 75 videos.
\item Sad - 71 videos.
\item Surprised - 70 videos.
\item Angry - 72 videos.
\item Disgust - 60 videos.
\item Fear - 60 videos.
\end{itemize}
Total Number of videos/identities: 484 videos/identities.
\\
Total number of frames: 250K approx.

\section{Preprocessing the dataset}
In this section we preprocess the dataset using detection and alignment and then annotate the frames/images based on their valence arousal score.
\subsection{Face Detection and Alignment}
The MTCNN\cite{zhang2016joint} or Multi-Task Cascaded Convolutional Neural Networks \cite{kollias6} was used for the face detection and the five facial landmarks was used for the alignment of faces.\\
This entire pipeline of Facial detection and alignment can be explained in the algorithm below:
\begin{algorithm}[H]
\caption{Facial Detection and Alignment}\label{alg:mtcnn}
\begin{algorithmic}[1]
\State We use a fully Convolutional Network\cite{dai2016r}, here called Proposal Network (P-Net), to obtain the candidate windows and their bounding box regression vectors in a similar manner as in \cite{farfade2015multi}. Then we use the estimated bounding box regression vectors to calibrate the candidates. After that, we use non-maximum suppression (NMS)\cite{neubeck2006efficient} to merge highly overlapped candidates.
\State All the candidates are then fed to another CNN, called the Refine Network (R-Net), which further rejects a large number of false candidates, performs calibration with bounding box regression, and NMS candidates merge.
\State This step is similar to the second step, but in this step we attempt to describe the face in more details. In particular, the network will output five facial landmarks’ positions which is basically used to align the faces.
\end{algorithmic}
\end{algorithm}
The three steps can also be explained as figure flow in the following page :
\begin{figure}[H]
\centering
\includegraphics[width=0.60\textwidth]{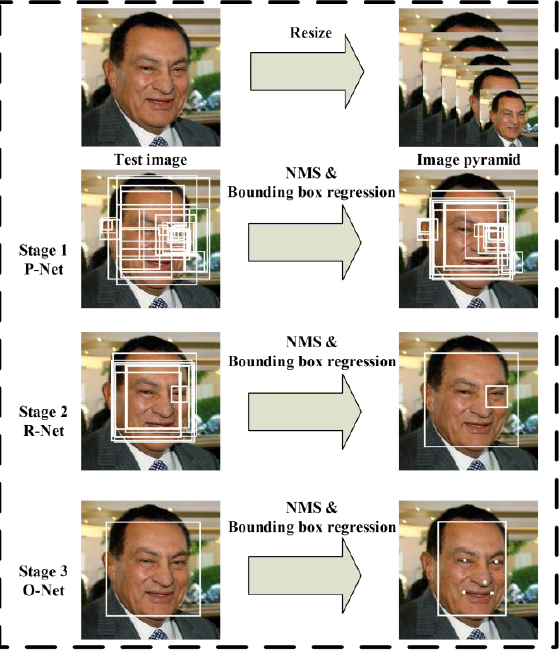}
\caption{The three step pipeline for Face Detection and Alignment}
\label{fig:mtcnn}
\end{figure}
\pagebreak
\subsection{Annotating the dataset}
In this subsection the valence arousal score was given to each frame/image based on its positive or negative range of emotions, as in \cite{kollias1,kollias2,kollias3}. We have defined the valence arousal score in the two dimensional space\cite{schubert1999measuring,kollias7,kollias14} with values ranging from -1 to +1 for valance and arousal each.\\
The following figure illustrates the metrics which we used to measure the extent of a particular emotion and whether the emotion is a positive emotion pr negative emotion.
\begin{figure}[H]
\centering
\includegraphics[width=0.65\textwidth]{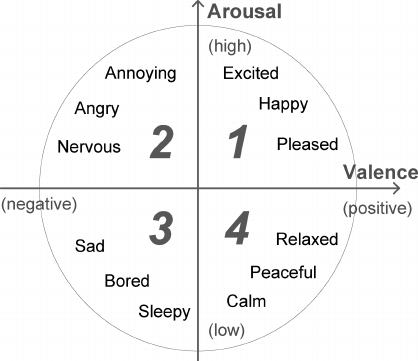}
\caption{Valence and Arousal metrics in 2D space}
\label{fig:va}
\end{figure}
\begin{flushleft}
\textbf{Valence score} is basically the measure of is positive or negative emotion of a particular person.\\
\end{flushleft}
\textbf{Arousal score} measures how calming or exciting the person is which means the intensity or extent of the particular emotion.\\
\\
If we consider an example:\\
For an angry image, the Valence score is $\sim -0.57$ and Arousal score is $\sim 0.63$\\ where the negative sign on valence score ensures a negative emotion and a high arousal score means he/she is very angry with a lot of excitement or energy on his/her face.
\\
Hence, this concludes the preprocessing part of the dataset. We collected a brand new dataset and preprocessed using detection, alignment and annotation.
\chapter{Method Implementation and Results}
In this chapter we use the existing StarGAN model and use it on the dataset we generated. We train the discriminator and generator with the images and finally test the images from our dataset on the trained model.
\section{Model Architecture}
We use the model architecture specified in the StarGAN paper\cite{choi2018stargan} and train the generator and the discriminator accordingly.\\
For the generator network, we use instance normalization\cite{ulyanov2016instance}
in all layers except the last output layer. For the discriminator network, we use Leaky ReLU\cite{xu2015empirical} with a slope of
-0.01.\\
The generator network architecture is given below:
\begin{figure}[H]
\centering
\includegraphics[width=1.1\textwidth]{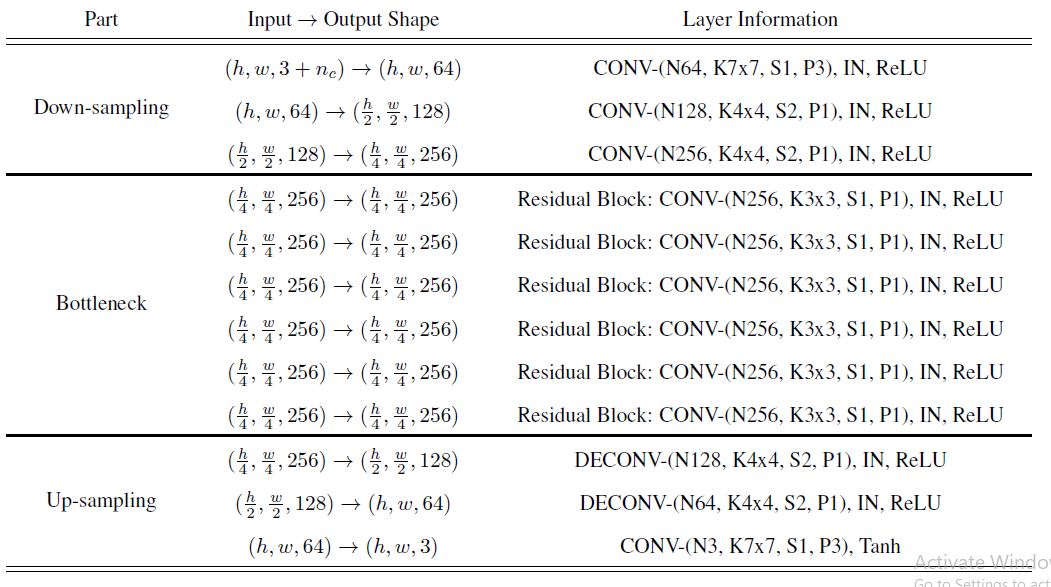}
\caption{Generator Network Architecture}
\label{fig:genArch}
\end{figure}
\pagebreak
The discriminator network architecture is given below:
\begin{figure}[H]
\centering
\includegraphics[width=1.1\textwidth]{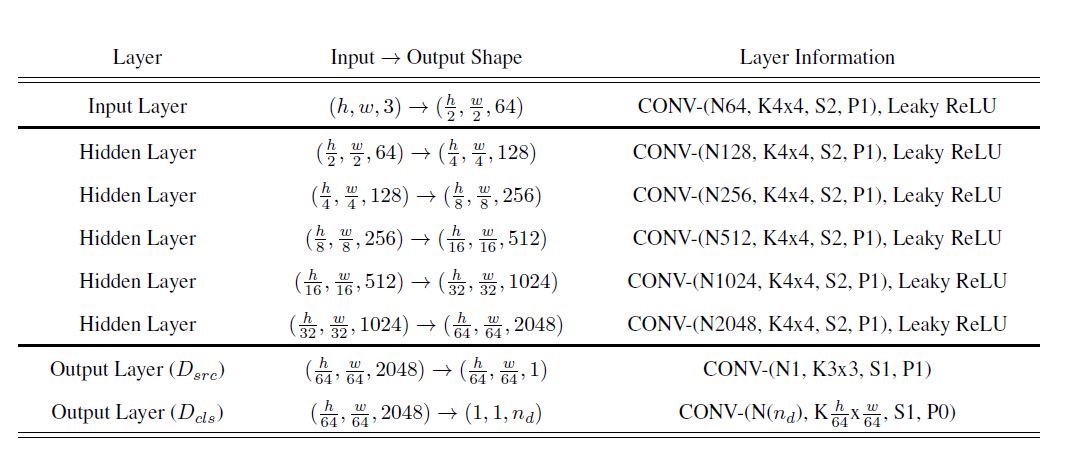}
\caption{Discriminator Network Architecture}
\label{fig:discArch}
\end{figure}
\vspace{-20mm}
\section{Training the dataset}
For \textbf{Generator}, Total number of parameters: 8436800.\\
For \textbf{Discriminator}, Total number of parameters: 44735424.\\
The models are trained using Adam \cite{kingma2014adam} with $\beta 1 = 0.5$ and $\beta 2 = 0.999$. We perform one generator update after five discriminator updates as in Wasserstein GANs\cite{gulrajani2017improved}. The batch size is set to 16 during training.\\
To produce higher quality images and improve the training process we generate the adversarial loss with the Wasserstein GANs\cite{gulrajani2017improved} objective with gradient penalty\cite{arjovsky2017wasserstein} which can be defined as:
\begin{equation}
    \begin{split}
        L_{adv} = \E_{x}[D_src(x)]-\E_{x,c}[D_{src}(G(x,c))]\\
        - \Lambda_{gp}E_{\hat{x}}[(\| \Delta_{\hat{x}}D_{src}(\hat{x})\| - 1)^{2}]
    \end{split}
\end{equation}
where $\hat{x}$ is sampled uniformly along a straight line between a pair of a real and a generated images.\\
After training the entire dataset we get the final training samples as illustrated in the following page:
\pagebreak
\begin{center}
\begin{tabular}{ c c c c c c c c}
 Input & Angry & Disgust & Fear & Happy & Neutral & Sad & Surprised
\end{tabular}
\end{center}
\vspace{-6mm}
\begin{figure}[H]
\centering
\includegraphics[width=0.75\textwidth]{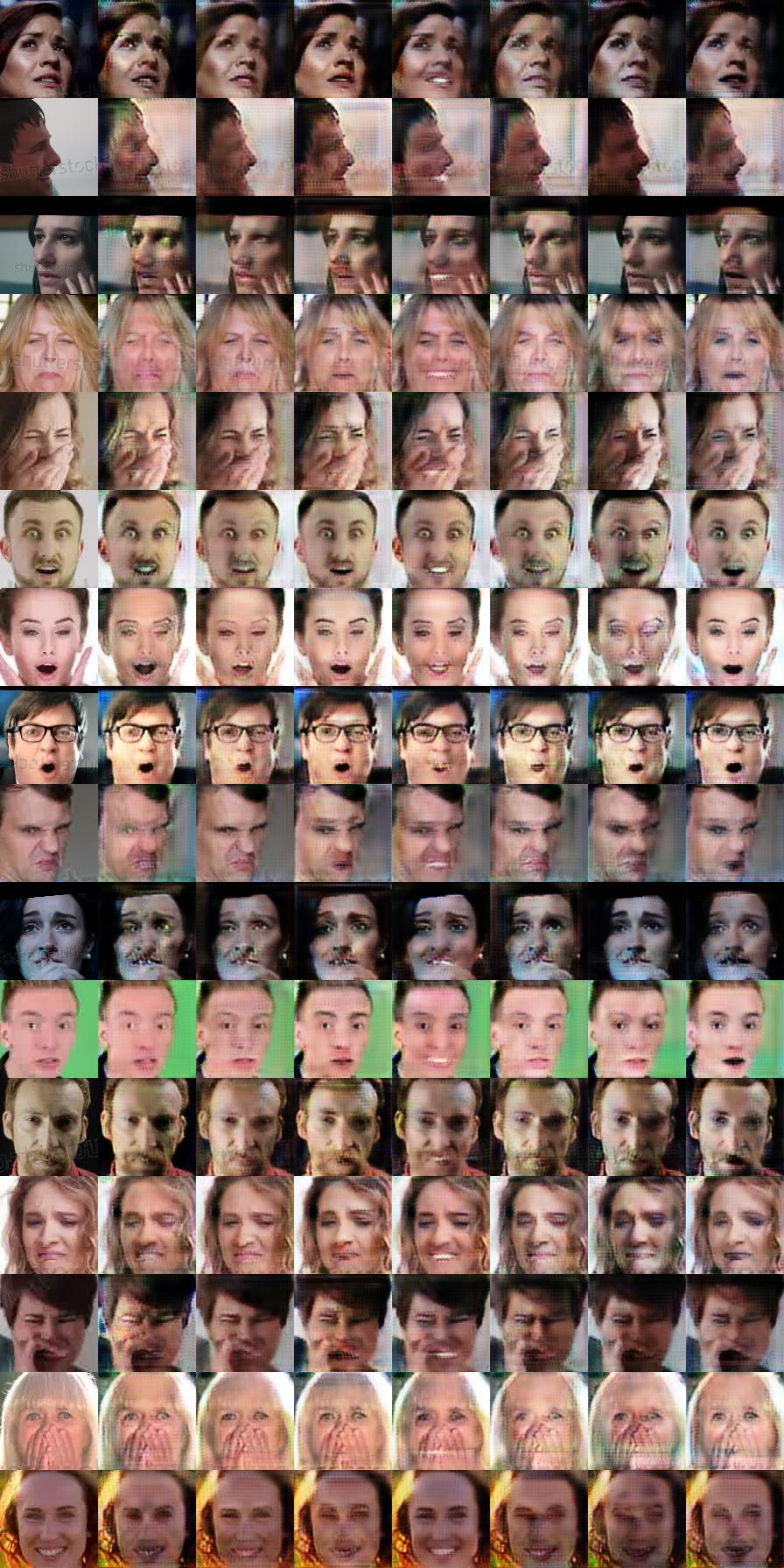}
\caption{Training Samples depicting the 7 different expressions}
\label{fig:train}
\end{figure}
\section{Testing the dataset}
After we have trained the entire dataset and generated the images and the 7 different expressions for the images we finally test the trained model using a few subjects to check its accuracy. We split the train and test data set as 90 percent and 10 percent respectively.\\

For subject 1
\vspace{-60mm}
\begin{center}
\begin{tabular}{ c c c c c c c c}
 Input & Angry & Disgust & Fear & Happy & Neutral & Sad & Surprised
\end{tabular}
\end{center}
\vspace{-87mm}
\begin{figure}[H]
\centering
\includegraphics[width=0.80\textwidth]{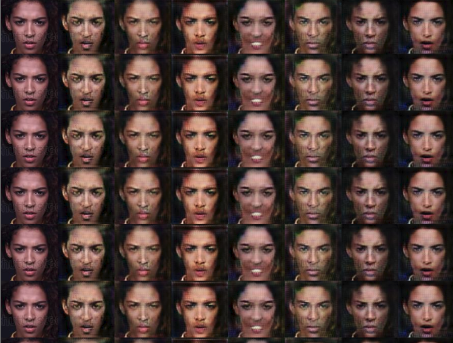}
\caption{Test Results for Subject 1}
\label{fig:rwoman}
\end{figure}
\pagebreak
For subject 2
\begin{center}
\begin{tabular}{ c c c c c c c c}
 Input & Angry & Disgust & Fear & Happy & Neutral & Sad & Surprised
\end{tabular}
\end{center}
\vspace{-6mm}
\begin{figure}[H]
\centering
\includegraphics[width=0.80\textwidth]{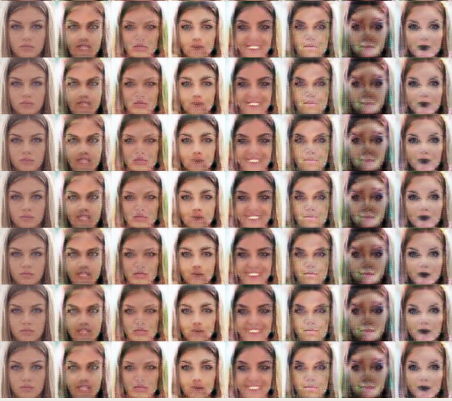}
\caption{Test Results for Subject 2}
\label{fig:rwoman2}
\end{figure}
So, as we can clearly see the trained model easily generates the 7 basic expressions for a given input image.\\
\textbf{Evaluation of the Discriminator} after 20000 iterations is:\\
Iteration [20000/20000],\\
D/loss\textunderscore real: -69.1152, D/loss\textunderscore fake: 22.4376, D/loss\textunderscore cls: 0.0002, D/loss\textunderscore gp: 1.2612,\\
\textbf{Evaluation of the Generator} after 20000 iterations is:\\
Iteration [20000/20000],\\
G/loss\textunderscore fake: -25.4440, G/loss \textunderscore rec: 0.1537, G/loss \textunderscore cls: 1.2868.
\chapter{Conclusion}
In conclusion, it can be clearly observed that the dataset gets classified into the 7 basic different expressions based on its valence arousal score.\\
We first preprocess the data using face detection, alignment and annotation; following which divide it into training and testing set. After that we feed the data set into the StarGAN model which then completes the training on 20000 iterations and around 250K frames/images. Finally, we test the pretrained model on some sample subjects and see how accurate the images get generated according to the valence arousal scores of each emotion \cite{kollias10}.\\
We can further improve the existing model by increasing the number of iterations which can produce even better accuracy on the test set.\\ We can also increase the number of subjects in the training datasets. For example we used around 500 different identities for training. This number can be further increased while dataset collection and preprocessing.

\bibliographystyle{unsrt}
\bibliography{bibs/sample}
\addcontentsline{toc}{chapter}{Bibliography}

\end{document}